\documentclass[a4paper]{article}
\usepackage[letterpaper]{geometry}
\usepackage{amsmath}
\usepackage{graphicx,color}
\usepackage{hyperref}
\usepackage{bbm, dsfont}
\usepackage{float}
\usepackage{cases}
\usepackage{algorithm}
\usepackage[noend]{algpseudocode}
\usepackage{longtable}
\usepackage{tabularx}
\usepackage{diagbox}
\usepackage{slashbox}
\usepackage{multirow}
\usepackage{booktabs}
\usepackage{siunitx}
\usepackage{subcaption}

\makeatletter
\def\BState{\State\hskip-\ALG@thistlm}

\usepackage{cite}
\usepackage{amssymb,amsfonts}
\usepackage{textcomp}
\usepackage{twocolceurws}

\title{Distributed Collapsed Gibbs Sampler for Dirichlet Process Mixture Models in Federated Learning}

\author{
Reda Khoufache \\
        {\small \textit{DAVID Lab}}\\ 
        {\small \textit{University of Versailles, Paris-Saclay University}}\\
        {\small Versailles, France}\\
        {\small reda.khoufache@uvsq.fr}
\and
Mustapha Lebbah \\
        {\small \textit{DAVID Lab}}\\ 
        {\small \textit{University of Versailles, Paris-Saclay University}}\\
        {\small Versailles, France}\\
        {\small mustapha.lebbah@uvsq.fr}
\and
Hanene Azzag \\
        {\small \textit{LIPN (CNRS UMR 7030)}}\\ 
        {\small \textit{Sorbonne Paris Nord University}}\\
        {\small Villetaneuse, France}\\
        {\small azzag@univ-paris13.fr}
\and
Etienne Goffinet \\
        {\small \textit{Technology Innovation Institute}}\\ 
        {\small Abu Dhabi, United Arab Emirates}\\
        {\small etienne.goffinet@tii.ae}
\and
Djamel Bouchaffra \\
        {\small \textit{CDAT}}\\ 
        {\small Algiers, Algeria}\\
        {\small djamel.bouchaffra@gmail.com}
}

\institution{}
\begin{document}
\maketitle
\begin{abstract}\small\baselineskip=10pt 
Dirichlet Process Mixture Models (DPMMs) are widely used to address clustering problems. Their main advantage lies in their ability to automatically estimate the number of clusters during the inference process through the Bayesian non-parametric framework. However, the inference becomes considerably slow as the dataset size increases. This paper proposes a new distributed Markov Chain Monte Carlo (MCMC) inference method for DPMMs (DisCGS) using sufficient statistics. Our approach uses the collapsed Gibbs sampler and is specifically designed to work on distributed data across independent and heterogeneous machines, which habilitates its use in horizontal federated learning. Our method achieves highly promising results and notable scalability. For instance, with a dataset of 100K data points, the centralized algorithm requires approximately 12 hours to complete 100 iterations while our approach achieves the same number of iterations in just 3 minutes, reducing the execution time by a factor of 200 without compromising clustering performance. The code source is publicly available at \url{https://github.com/redakhoufache/DisCGS}.
\end{abstract}

\section{Introduction}
Clustering is an unsupervised learning method that aims to partition data into clusters such that elements of the same cluster are similar while those of different clusters are dissimilar. It plays a crucial role in various domains, such as data analysis, pattern recognition, and data mining. To address clustering problems, mixture models have emerged as popular generative probabilistic models. They assume the existence of a mixture of distributions over the observation space, where each cluster is associated with a latent component distribution. The flexibility of mixture models allows researchers and practitioners to analyze and interpret data effectively.

Dirichlet Process Mixture Models (DPMMs) represent an extension of mixture models into Bayesian non-parametric models. DPMMs assume a mixture model with an infinite number of latent components and make a prior distribution over the model parameters. In literature, two common processes are used to represent the Dirichlet Process (DP). The first is the Chinese Restaurant Process (CRP) \cite{10.1214/aos/1176342871}, which induces a distribution over the space of partitions. The second representation is known as the Stick-Breaking (SB) process \cite{10.2307/24305538}, which provides a construction of the Dirichlet Process distribution. In \cite{https://doi.org/10.48550/arxiv.1801.00513}, the authors proved that each of these representations could be derived from the other. 


The Gibbs Sampling algorithm \cite{10.2307/1390653} is a famous MCMC method that performs the inference of DPMM's parameters. This algorithm iteratively updates the membership vector and the parameters associated with each cluster. When the prior conjugacy assumption holds, it becomes possible to integrate out the parameters. Consequently, the inference process only requires sampling the memberships, leading to a variant known as collapsed Gibbs sampling.

While the DPMM has the advantage of automatically estimating the number of clusters and discovering new structures, its inference process becomes prohibitively slow and non-scalable when the number of observations is large. This poses a significant inconvenience for handling massive datasets and limits the applicability of the DPMM in real-world scenarios.

Distributed computing consists in distributing data across multiple workers; the 
workers could be components in the edge network, cores in a 
cluster, computers in a network, processors in a computer, etc. This distribution 
enables them to execute parallel computations independently. It presents an opportunity 
to accelerate computations and circumvent memory limitations, making it 
particularly suitable for handling large datasets. 

Federated Learning (FL), a pioneering concept introduced by 
\cite{konecny2016federated}, 
revolutionized large-scale distributed learning by harnessing the machine intelligence 
residing in local devices.
Federated learning is used to train machine learning models directly on edge devices or computing nodes. 
In the following, we will refer to a component of the edge network or a network node as a "worker".  One of the defining characteristics of FL is that the data distributions may possess very different properties across the workers. Hence, any potential FL clustering method is explicitly required to be able to work under the heterogeneous data setting. Therefore, the availability of DPMM models tailored to these distributions becomes an indispensable requirement in our case. However, due to its highly decentralized nature, FL poses numerous statistical and computational challenges. In addition, the complexity of the clustering problem increases significantly when applied in a federated context. Furthermore, in many real-world applications, data is actually stored at the edge, spanning nodes, phones, and computers.

In this paper, we address a specific DPMM problem that arises in federated learning, contributing to the advancement of this exciting field. We propose a new distributed MCMC-based approach for DPMM inference (DisCGS) using sufficient statistics. It is important to emphasize that our approach focuses on a specific inference algorithm for the DPMM, the Collapsed Gibbs Sampler proposed in \cite{10.2307/1390653}. The scalability of this algorithm has not been addressed yet.

\begin{figure}[ht]
    \centering
    \includegraphics[scale=0.27]{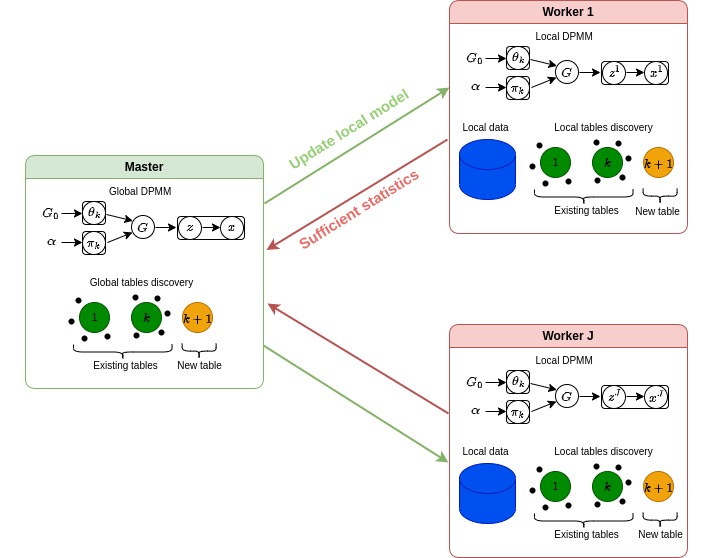}
    \caption{DisCGS workflow. Workers have only access to their own local data. Each worker infers a local DPMM and discovers local tables (clusters). Sufficient statistics are sent from each worker to the master, which infers the global DPMM and estimates the global partition. The global model is shared to each worker.}
    \label{fig:gnwrfl}
\end{figure}

We highlight our key contributions as follows:
{\bf(1)} In the federated context, we have developed a new formulation of distributed MCMC inference of the DPMM using the Master/Worker architecture.
The data is evenly distributed among the workers, ensuring a balanced workload. The workers operate independently and do not share information with each other. Instead, they communicate solely with the master, exchanging minimal information. {\bf(2)} Each worker executes a local collapsed Gibbs sampler, which enables the discovery of local clusters and the inference of a local DPMM. Then, the sufficient statistics associated with each local cluster are sent to the master.
{\bf(3)} At the master level, a global collapsed Gibbs sampler is performed using only sufficient statistics. This allows the synchronization and estimation of the global DPMM and global clustering structure without accessing the content of each cluster. Theoretical background and computational details are provided. The general workflow of our model is illustrated in Figure \ref{fig:gnwrfl}.
{\bf(4)} We conduct several experiments to validate our method's merits and scalability with the number of nodes on large datasets. For a dataset of 100K data points, the centralized collapsed Gibbs sampler 
takes 12 hours to perform 100 iterations, whereas our approach 
does in just 3 minutes. 
{\bf(5)} Distributing the collapsed Gibbs sampler for DPMMs is a crucial intermediate step towards the distribution of Bayesian non-parametric coclustering \cite{Meeds07nonparametricbayesian} and Multiple coclustering \cite{goffinet2021multicoclustering}. 


\section{Related Work}
\indent Clustering problems in federated learning have drawn attention in recent studies. For instance, \cite{9499980} introduced an approach for federated K-Means clustering, \cite{DBLP:journals/corr/abs-2201-07316} proposed a Federated Fuzzy $c$-Means algorithm. In \cite{fedGen}, the authors presented a federated clustering method using probabilistic models. 

Several parallelized and distributed approaches have been proposed to address the slow inference of DPMMs. \cite{lovell} introduced a reparameterization of the Dirichlet Process that facilitates the learning of clusters representing the data and super-clusters defining the granularity of parallelization. \cite{pmlr-v28-williamson13} incorporated auxiliary variables into the Dirichlet Process and Hierarchical Dirichlet Process to establish a conditional independence structure, enabling a parallel Gibbs sampler without the need for approximations. However, \cite{pmlr-v32-gal14} demonstrated that these approaches are impractical due to extremely unbalanced data distributions.
Another parallelized MCMC sampler was proposed by \cite{NIPS2013_bca82e41}, which combines a non-ergodic, restricted Gibbs sampler in the workers with split/merge proposals \cite{10.2307/1391150} at the master level to ensure ergodicity. However, this method is implemented only for a single machine. This approach is extended in \cite{dinari2019distributed} to a distributed version across multiple machines using a distributed-memory model.

In \cite{pmlr-v37-gea15}, the authors proposed a distributed inference algorithm for DPMMs based on a slice sampler developed by \cite{doi:10.1080/03610910601096262}. They addressed the scalability by developing a distributed architecture that estimates the Dirichlet Process and Hierarchical Dirichlet Process mixture models. \cite{Wang2017ScalableEO} presented a scalable estimation method for DPMMs in a distributed architecture. Their approach allows the creation of new components locally in individual computing nodes and proposes a probabilistic consolidation scheme for merging the created components. In the \cite{inproceedings}, the authors introduced another distributed Markov Chain Monte Carlo (MCMC) inference method based on the Gibbs sampler. Their approach performs a local Gibbs sampler to infer new local clusters and a new Gibbs sampler on the means of each cluster to synchronize the clustering. However, this method assumes a Dirichlet Process Gaussian mixture model with known variances. This assumption emphasizes known variances within and between clusters due to the prior, which limits the method's applicability. 

The concept of using sufficient statistics is well-known and has been widely used to distribute the DPMM inference in many works, including \cite{meguelati:hal-01999453, dinari2019distributed, Wang2017ScalableEO, pmlr-v37-gea15}. However, it has not been applied to distribute the Collapsed Gibbs Sampler, which constitutes the main focus of this paper. Moreover, none of the existing methods is exploitable to enhance the scalability and the distribution of Bayesian non-parametric coclustering \cite{Meeds07nonparametricbayesian} and Multiple coclustering \cite{goffinet2021multicoclustering}. Moreover, to the best of our knowledge, no prior work has been proposed on federated clustering using Dirichlet Process models.

\section{Background}
\subsubsection*{Model definition}
Let $n$ and $d$ be two positive integers, $\mathbf{x}=(x_1,\cdots, x_n)^T \in \mathbb{R}^{n\times d}$ the observed data, where $(\cdot)^T$ denotes the transpose operator. Let $\mathbf{z} = (z_1, \cdots, z_n)$ be the membership vector, where $z_i$ is a latent variable such that $z_i = k$ means that the observation $x_i$ belongs to the cluster $k$. The DPMM assumes that the observed data are generated according to the following model:
$$
\begin{gathered}
x_{i} \mid\left\{z_{i}=k, \theta_{k}\right\} \stackrel{\text { i.i.d. }}{\sim} f\left(x_i,\theta_{k}\right),\, \forall i\in\{1,\cdots, n\} \\
\theta_{k} \stackrel{\text { i.i.d. }}{\sim} G_{0},\, \forall k\in\{1,2,\cdots\} \\
\,z_{i} \stackrel{\text { i.i.d. }}{\sim} \operatorname{Mult}\left(\pi\right),\, \forall i\in\{1,\cdots, n\}\\
\pi \sim SB(\alpha).
\end{gathered}
$$
Under this model assumption, an observation $x_i$ is generated by first drawing $z_i$ from the Multinomial distribution parameterized by the vector of weights $\pi = (\pi_k)_{k=1}^\infty$ (also called the mixture proportions), and then $x_i$ is sampled from $f(x_i,\theta_{z_i})$, where $f(\cdot,\theta_k)$ is the parameterized component distribution associated to the cluster $k$. The components parameters $\theta_k$ follow a prior distribution $G_0$ (also called the base distribution); in the multivariate Gaussian case, we have $\theta_k = (\mu_k,\Sigma_k)$. We let $\Theta = \{\theta_k,\, k>1\} $ be the set of components parameters. The mixture proportions follow the Stick-Breaking process \cite{10.2307/24305538} parameterized by a concentration parameter $\alpha>0$. We recall that the Stick-Breaking process is defined as follows: 
$$
\pi_{k}=v_{k} \prod_{k^{\prime}=1}^{k-1}\left(1-v_{k^{\prime}}\right), \,\,\,\, v_{k} \stackrel{\text { i.i.d. }}{\sim} \operatorname{Beta}\left(1, \alpha \right).
$$
We have $\pi_k>0$ for every $k>1$ and $\sum_{k=1}^{\infty} \pi_k = 1$. It is proved by \cite{10.2307/24305538} that the distribution $G = \sum_{k=1}^{\infty} \pi_k \delta_{\theta_k}$ (where $\theta_k$ are sampled i.i.d according to $G_0$ and $\delta$ is the indicator function) follows the Dirichlet Process with a concentration parameter $\alpha$ and a base distribution $G_0$, noted $DP(\alpha, G_0)$. In the following, we denote $\Omega = (\alpha, G_0)$ the hyper-parameter set. 
\subsubsection*{Inference.}\label{sec inf}
The Gibbs Sampling algorithm \cite{10.2307/1390653} is a popular algorithm used for the DPMM inference based on Monte-Carlo sampling. It alternates between updating the membership vector and updating the parameters associated with each cluster. In the membership update step, the Gibbs sampling simulates the posterior distribution $\mathrm{p}(\mathbf{z}\mid\mathbf{x},\Theta, \Omega)$ by sampling each $z_i$ from the conditional distribution $\mathrm{p}(z_i \mid \mathbf{x}, \mathbf{z}_{-i},\Theta, \Omega)$, where  $\mathbf{z}_{-i} = \{z_{l},\,\, l\neq i\}$ is the set of the remaining memberships. In the parameter update step, the component parameter of each cluster $k$ is updated by sampling according to the posterior distribution $\mathrm{p}(\theta_{k}| \mathbf{x}_k, G_0)$, with $\mathbf{x}_k = \{x_i,\, z_i = k\}$ the set of observations that belong to cluster $k$. The computing of such distributions is analytically tractable when $G_0$ is the conjugate prior to the density $f$. 

The collapsed Gibbs sampler, which corresponds to the third algorithm proposed in \cite{10.2307/1390653} skips the parameter sampling step because under the conjugacy assumption, it is possible to integrate out the parameters $\theta_k$ which allows to directly compute the predictive (prior and posterior) without the parameter's values. Each membership $z_i$ is sampled according to $\mathrm{p}(z_i\mid \mathbf{z}_{-i}, \mathbf{x}, \Omega) \propto$: 
\begin{numcases}{}
 n_{k} \mathrm{p}(x_i \mid z_i = k, \mathbf{x}_k, G_0), & \text{existing cluster} $k$, \label{eq 1} \\
 \alpha \mathrm{p}(x_i \mid \Omega), & \text {new cluster,} \label{eq 2}
\end{numcases}
where $n_k$ is the cardinal of cluster $k$. The posterior predictive $\mathrm{p}(x_i \mid z_i = k,\mathbf{x}_k, G_0)$ and prior predictive $\mathrm{p}(x_i\mid \Omega)$ can be obtained by integrating over $\theta$. 

In the multivariate Gaussian case with unknown mean and variance, we choose the Normal 
 Inverse Wishart \cite{gelman1995bayesian} (NIW) to be the conjugate prior $G_0$, with hyper-parameters $\lambda_0 = (\mu_0,\kappa_0,\Psi_0,\nu_0)$, where $\mu_0 \in \mathbb{R}^d$ is the 
 prior mean, $\kappa_0 \in \mathbb{R}$ is the number of prior measurements, $\nu_0$ is the 
 degrees of freedom and $\Psi_0\in \mathbb{R}^{d\times d}$ is the precision matrix. In this case, the probabilities in equations \ref{eq 1} and \ref{eq 2} can be computed analytically \cite{murphy2007conjugate}. 


To summarize, the collapsed Gibbs sampler is an efficient MCMC method because it avoids component parameter sampling. This technique, known as Rao-Blackwellization, is due to the Rao-Blackwell theorem \cite{10.1214/aoms/1177730497}. This theorem ensures that the estimator's variance obtained by integrating out $\theta$ is always lower than or equal to the one of a direct estimator. This theorem remains true in collapsed Gibbs sampling \cite{b084eadd-5e75-3a50-a017-472b61e6a004}. 


\section{The proposed method}
The collapsed version of the Gibbs sampling inference inspires our approach. The main objective of DisCGS is to make the inference scalable while keeping the MCMC's precision and the DPMM's flexibility. Our approach consists in distributing the data evenly over the workers and alternating at each iteration between two steps: Worker level and Master level.
%
Below, we provide a theoretical detailed 
of the collapsed Gibbs samplers executed at each level. The workflow of our approach is described in figure \ref{fig:gnwrfl}

\subsection{DisCGS at worker level}\label{sec worker_level}
We denote by $\mathbf{x}^j = \{x_1^j,\cdots, x_{n^j}^j\}$ the set of observations assigned to the $j$-th worker, $n^j$ the cardinal of $\mathbf{x}^j$,  and $\mathbf{z}^j = \{z_1^j,\cdots z_{n^j}^j\}$ the local membership vector such that $z_i^j = k$ means that the observation $x_i^j$ (assigned to the worker $j$) belongs to the local cluster $k$. 

In this level, the local memberships are updated one by one using the collapsed version of Gibbs sampling detailed in section \ref{sec inf}. Each $z_i^j$ is updated by sampling from $\mathrm{p}(z_i^j\mid \mathbf{z}^j_{-i}, \mathbf{x}^j, \Omega) \propto$
\begin{numcases}{}
n^j_{k} \mathrm{p}(x^j_i \mid z^j_i = k, \mathbf{x}^j_k, G_0), & \text{existing cluster } $k$, \label{eq 8} \\
 \alpha \mathrm{p}(x^j_i \mid \Omega), & \text {new cluster,} \label{eq 9}
\end{numcases}
where $n^j_{k}$ is the size of cluster $k$ in worker $j$ and $\mathbf{x}_k^j$ is the contents of cluster $k$ in worker $j$.  The posterior and prior predictive distributions are computed analytically using the same global prior $G_0$, as detailed in \cite{murphy2007conjugate}.  After updating the local membership vector, we compute the sufficient statistics \cite{silvey2017statistical} associated with each cluster. In the multivariate Gaussian case, the sufficient statistics $(T_{k}^{j},S_{k}^{j})$ for a cluster $\mathbf{x}_{k}^{j}$ are given by \cite{gelman1995bayesian}:
\begin{align}
T_k^j &= \frac{1}{n_k^j} \sum_{x\in\mathbf{x}_k^j} x \in \mathbb{R}^d,\label{eq 10}\\
S_k^j &= \sum_{x\in \mathbf{x}_k^j} (x - T_k^j)(x - T_k^j)^T \in \mathbb{R}^{d\times d}.\label{eq 10bis}
\end{align}

Finally, sufficient statistics and the sizes of each cluster are sent to the master. The DisCGS inference process at the worker level is described in Algorithm \ref{DPMWL}.  

\begin{algorithm}
\caption{DisCGS inference at worker level}\label{DPMWL}
\begin{algorithmic}[1]
\State \textbf{Input}: Dataset $\mathbf{x}^j$, concentration parameter $\alpha$ and prior $G_0$.
\BState \textbf{For} $i\leftarrow 1$ \textbf{to} $n^j$ \textbf{do}:
\BState\indent Remove $x_i^j$ from its local cluster.
\BState\indent Compute $\mathrm{p}(z_i^j\mid \mathbf{z}^j_{-i}, \mathbf{x}^j, \Omega)$ as defined by eq. \ref{eq 8} and \ref{eq 9}.
\BState\indent Sample $z_i^j$.
\BState\indent Add $x_i^j$ to its new cluster.
\BState \textbf{For} $k\leftarrow 1$ \textbf{to} $K$ \textbf{do}:
\BState \indent Compute the sufficient statistics $(T_k^j,S^j_k)$ using eq. \ref{eq 10} and \ref{eq 10bis}.
\State \textbf{Output}: Sufficient statistics $\{(T_1^j,S_1^j)\cdots,(T_K^j,S_K^j)\}$ and cluster sizes $(n_1^j,\cdots,n_K^j)$.
\end{algorithmic}
\end{algorithm}

\subsection{DisCGS at master level}\label{sec master_level}

The master receives from each worker the sample size of each cluster and its associated sufficient statistics. The objective is to estimate the membership vector $\mathbf{z}=(z_1,\cdots,z_n)$ and update the prior hyper-parameters of each cluster. 

In this level, the observations are assigned by batch; a batch corresponds to a set of observations that belong to the same cluster.  In fact, instead of assigning the observations one by one to their clusters, we assign a group of observations that already share the same local cluster (i.e., at the worker level) to a global cluster at the master level. Hence, the observations assigned to the same global cluster will share the same label. We sample the global membership $z_h^j$ of the cluster $\mathbf{x}_h^j$ (the local cluster $h$ of worker $j$) according to $\mathrm{p}(z_h^j\mid\mathbf{z}^{-j}_{-h}, \mathbf{x}, \Omega) \propto$
\begin{numcases}{}
n_{k} \mathrm{p}(\mathbf{x}^j_h \mid z^j_h = k, \mathbf{x}_k, G_0), & \text{existing cluster }k, \label{eq 11} \\
 \alpha \mathrm{p}(\mathbf{x}^j_h \mid G_0), & \text {new cluster,} \label{eq 12}
\end{numcases}

In practice, the joint posterior predictive and the joint prior predictive distributions (equations \ref{eq 11} en \ref{eq 12} respectively) are computed analytically by only using sufficient statistics, i.e. without having access to the content of cluster $\mathbf{x}_h^j$. In fact, we have: 
$$
\mathrm{p}\left(\mathbf{x}_{h}^{j} \mid \Omega\right)=\pi^{-n_{h}^{j}\frac{d}{2}}\cdot \frac{\kappa_0^{d / 2}}{\left(\kappa_{h}^{j}\right)^{d / 2}} \cdot \frac{\Gamma_d\left(\nu_{h}^{j} / 2\right)}{\Gamma_d\left(\nu_0 / 2\right)} \cdot \frac{\left|\Psi_0\right|^{\nu_0 / 2}}{\left|\Psi_{h}^{j}\right|^{\nu_{h}^{j} / 2}}
$$
where $|\cdot|$ is the determinant, and the hyper-parameter values $(\mu_h^j, \kappa_h^j, \Psi_h^j, \nu_h^j)$ are obtained as follows:
$$
\begin{gathered}
\mu_{h}^{j}=\frac{\kappa_0 \mu_0+n_{h}^{j} T_{h}^{j}}{\kappa_{h}^{j}}, \quad \kappa_{h}^{j}=\kappa_0+n_{h}^ {j}, \quad \nu_{h}^{j}=\nu_0+n_{h}^{j}, \\
\Psi_{h}^{j}=\Psi_0+S_h^j+\frac{\kappa_0 n_{h}^{j}}{\kappa_{h}^{j}}\left(\mu_0-T_{h}^{j}\right)\left(\mu_0-T_{h}^{j}\right)^T,
\end{gathered}
$$
where $T_h^j$ and $S_h^j$ are the sufficient statistics obtained from the workers. Moreover, we have $\mathrm{p}(\mathbf{x}^j_h \mid z^j_h = k, \mathbf{x}_k, G_0)=$ 
$$
\pi^{\frac{-dn_{h}^{j}}{2}}\cdot \frac{\kappa_k^{d / 2}}{\left(\kappa_{h}^{j}\right)^{d / 2}} \cdot \frac{\Gamma_d\left(\nu_{h}^{j} / 2\right)}{\Gamma_d\left(\nu_k / 2\right)} \cdot \frac{\left|\Psi_k\right|^{\nu_k / 2}}{\left|\Psi_{h}^{j}\right|^{\nu_{h}^{j} / 2}}
$$
and the posterior distribution parameters $(\mu_k, \kappa_k, \Psi_k, \nu_k)$ associated to the global cluster $k$, updated from the prior:
$$
\begin{gathered}
\mu_{k}=\frac{\kappa_0 \mu_0+n_{k} T_k}{\kappa_{k}}, \quad \kappa_{k}=\kappa_0+n_{k}, \quad \nu_{k}=\nu_0+n_{k}, \\
\Psi_{k}=\Psi_0+S_k+\frac{\kappa_0 n_{k}}{\kappa_{k}}\left(\mu_0-T_k\right)\left(\mu_0-T_k\right)^T.
\end{gathered}
$$

With $T_k$ and $S_k$, the aggregated sufficient statistics obtained when local clusters are assigned to the same global cluster $k$ are computed as follows: 
$$
\begin{gathered}
    T_k = \frac{1}{n_k} \sum_{j,h|\,\mathbf{z_h^j=k}} n_h^j\cdot T_h^j,\\ S_k = \sum_{j,h|\,\mathbf{z_h^j=k}} S_h^j + \sum_{j,h|\,\mathbf{z_h^j=k}}\left( n_j^h \cdot T_h^j\cdot {T_{h}^{j}}^T \right) - n_k \cdot T_k\cdot T_k^T.
\end{gathered}
$$
 The inference process at the master level is
described in Algorithm \ref{DPMML}. 

\begin{algorithm}
\caption{DisCGS inference at master level}\label{DPMML}
\begin{algorithmic}[1]
\State \textbf{Input}: Sufficient statistics, cluster sizes, $\alpha$ and prior $G_0$. 
\BState \textbf{For} each $(j,h)$ \textbf{do}:
\BState\indent Remove $\mathbf{x}_h^j$ from its global cluster.
\BState\indent Compute $\mathrm{p}(z_h^j\mid \mathbf{z}^{-h}_{-i}, \mathbf{x}_h^j, \Omega) \propto$ as defined by eq. \ref{eq 11} and \ref{eq 12}.
\BState\indent Sample $z_h^j$.
\BState\indent Add $\mathbf{x}_h^j$ to its new global cluster.
\BState Update the membership vector $\mathbf{z}$.
\State \textbf{Output}: Membership vector $\mathbf{z}$.
\end{algorithmic}
\end{algorithm}

\subsection{Collapsed Gibbs sampler in federated learning}

In federated learning, the observations are distributed on different workers "components". This corresponds to the horizontal decomposition of the dataset. In this context, each worker is initialized with the same global model, which is identical to the model present on the server. Then, each worker updates its model using its private data through the collapsed Gibbs sampler detailed in section \ref{sec worker_level}; this step allows the estimation of the local clusters and the local model. Then, the sufficient statistics and cluster sizes associated with each local cluster are computed and transmitted to the master "server". The master proceeds to update the global model and to estimate the global clustering structure without having access to the data. This process is achieved using the collapsed Gibbs sampler detailed in section \ref{sec master_level}. These updates are then shared with each worker, which allows them to update the local model with the global model. This iterative process continues alternating between the worker and master steps until the global model is fully estimated.

\section{Experiments}

To evaluate the effectiveness of our approach, we conduct three types of experiments on synthetic and real-world datasets. Firstly, we compare our distributed algorithm with other state-of-the-art clustering algorithms in terms of clustering performance and convergence rate. Secondly, we compare the execution time and clustering performance of our distributed algorithm DisCGS and the centralized CGS (Collapsed Gibbs Sampler from \cite{10.2307/1390653}) on synthetic datasets of different sizes. Lastly, we investigate the scalability of DisCGS by increasing the number of nodes while keeping the number of observations fixed. For this purpose, we execute our distributed algorithm on $10^6$ data points multiple times, varying the number of cores.  

\subsection{Implementation settings and distributed environment}

In the following experiments, we use an uninformative prior $\textrm{NIW}$ for both CGS and DisCGS algorithms. Therefore, both methods are implemented by setting the $\textrm{NIW}$ hyper-parameters as follows: $\mu_0$ and the matrix precision $\Psi_0$ are respectively set to be empirical mean vector and covariance matrix of all data, as we want them as uninformative as 
possible. $\kappa_0$ and $\nu_0$ represents our confidence in $\mu_0$ and $\Psi_0$, are set to their lowest 
values, which are $1$ and $d+1$, respectively, where $d$ is the dimension of the observation space. The initial state is a one-cluster 
partition. Finally, we have executed the distributed algorithms 
using the Neowise machine (1 CPU AMD EPYC 7642, 48 cores/CPU) hosted by the cluster grid5000 
\footnote{\url{https://www.grid5000.fr/w/Grid5000:Home}}. The centralized algorithm is executed on the same machine using one 
core.

\subsection{Clustering performance}
To evaluate the clustering performance of our algorithm, we compare our approach with two distributed algorithms for the DPMM inference:  M-R\footnote{The code source is taken from: \url{https://github.com/wangruohui/distributed-dpmm}}\cite{pmlr-v37-gea15} and SubC\footnote{The code source is taken from: \url{https://github.com/BGU-CS-VIL/DPMMSubClusters.jl/tree/master}} \cite{dinari2019distributed}, and two parallelized clustering algorithms: Kmeans and GMM (for both methods, we have used the Spark Mllib implementation). These two parametric methods require the number of clusters. To ensure a fair comparison of the clustering quality with our approach, we set the number of clusters in these methods equal to the one inferred by DisCGS. It is important to mention that SubC and M-R are the only approaches of distributed inference for DPMM for which open-source working code is available. 

All the executions are performed on the Neowise machine by distributing the data evenly on $32$ cores. In this experiment, we use $8$ different datasets described in table \ref{tab:datasetsdesc}. Due to the smaller size of the EngyTime dataset, we distributed the data only on two workers. The image datasets are encoded using a variational auto-encoder, and each image is encoded into an $8$-dimensional vector.

To evaluate the clustering performance, we compute the three clustering metrics: Adjusted Rand Index (ARI) \cite{hubert1985comparing}, Normalized Mutual Information (NMI)\cite{nmi}, and clustering accuracy (ACC) \cite{acc}.

\begin{table*}[t!]
\centering
\begin{tabular}{llcll}
\toprule
{Dataset} & {$n$} & {$d$} & {$K$} & {Description}  \\  
\midrule
{EngyTime}     & 4096 & 2 & 2 &  {Synthetic dataset generated from two Gaussian mixtures.} \\ 

{Synthetic 10K}     & 10000 & 2 & 6 &  {Synthetic dataset generated from Gaussian components of dimension $2$.} \\ 
{Mnist}         & 70000   & 8 & 10 &  {Handwritten digits, Lecun et al.} \\
{Fashion mnist} & 70000   & 8 & 10 &  {Zalando's article images \cite{xiao2017fashionmnist}.} \\
{Letter}        & 103600  & 8 & 26 &  {Handwritten letters \cite{7966217}.} \\
{Balanced}      & 131600  & 8 & 47 &  {Both of handwritten digits and letters \cite{7966217}.} \\
{Digits}        & 280000  & 8 & 10 &  {Handwritten digits \cite{7966217}.} \\    
{UrbanGB}       & 360177  & 3 & 469 &  {Coordinates (longitude and latitude) of road accidents \cite{Dua:2019}.} \\    
 \bottomrule
  \end{tabular}
      \caption{The description of the datasets used to evaluate the clustering performance of our distributed inference approach. $n$ denotes the size, $d$ the dimension, $K$ the true number of clusters.}
      \label{tab:datasetsdesc}
\end{table*}


Table \ref{clustresults} reports the mean and the standard deviation of the clustering metrics, ARI, NMI, and ACC, achieved by each method on each dataset over $10$ trials. The results show that our proposed method (DisCGS) outperforms other methods or achieved the second-best 
score in almost all the datasets for the three clustering metrics. It is important to note that in this experiment, we only focus on comparing 
the clustering performance of the different methods; we do not compare the execution times of the three methods because the other approaches 
proposed an inference algorithm that differs from the collapsed Gibbs sampler.
\begin{table}[t!]
\resizebox{\columnwidth}{!}{%
  \begin{tabular}{llccccc}
    \toprule
    {Dataset}                  &       & {DisCGS} & {M-R} &  {SubC} & {GMM} & {Kmeans} \\
    \midrule
       \multirow{2}{*}{EngyTime}       & ARI & $\mathbf{0.94} \mp 0.07 $ & $0.47 \mp 0.04 $ & $\underline{0.87} \mp 0.00$  & $0.73 \mp 0.13$ & $0.54 \mp 0.20$ \\
                                       & NMI & $\mathbf{0.92} \mp 0.07 $ & $0.42 \mp 0.02 $ & $\underline{0.79} \mp 0.00 $ & $0.67 \mp 0.09$ & $0.56 \mp 0.11$ \\
                                       & ACC & $\underline{0.96} \mp 0.04 $ & $0.75 \mp 0.04 $ & $\mathbf{0.97} \mp 0.00 $ & $0.85 \mp 0.11$ & $0.68 \mp 0.20$ \\ 
      \midrule
        \multirow{2}{*}{Synthetic 10K} & ARI & $\mathbf{0.80} \mp 0.06$ & $0.10 \mp 0.01 $ & $0.38 \mp 0.07 $  & $\underline{0.80} \mp 0.07$ & $0.75 \mp 0.03$\\
                                       & NMI & $\underline{0.86} \mp 0.04$ & $0.12 \mp 0.01 $ & $0.61 \mp 0.08 $  & $\mathbf{0.88} \mp 0.04$ & $0.85 \mp 0.01$\\ 
                                       & ACC & $\mathbf{0.88} \mp 0.06$ & $0.29 \mp 0.01 $ & $0.38 \mp 0.08 $  & $\underline{0.85} \mp 0.08$ & $0.76 \mp 0.05$\\ 
      \midrule
        \multirow{2}{*}{Mnist}         & ARI & $\mathbf{0.72} \mp 0.01$ & $0.20 \mp 0.07 $ & $\underline{0.66} \mp 0.04 $  & $0.39 \mp 0.02$ & $0.26 \mp 0.01$\\
                                       & NMI & $\underline{0.74} \mp 0.00$ & $0.38 \mp 0.08 $ & $\mathbf{0.79} \mp 0.01$  & $0.69 \mp 0.01$ & $0.62 \mp 0.00$\\ 
                                       & ACC & $\mathbf{0.79} \mp 0.01 $ & $0.30 \mp 0.06 $ & $\underline{0.71} \mp 0.03 $ & $0.38 \mp 0.02$ & $0.23 \mp 0.01$\\ 
      \midrule
        \multirow{2}{*}{Fashion-Mnist} & ARI & $\mathbf{0.45} \mp 0.02 $ & $0.35 \mp 0.02 $ & $0.40 \mp 0.02 $        & $\underline{0.41} \mp 0.02$ & $0.37 \mp 0.02$\\
                                       & NMI & $\mathbf{0.60} \mp 0.01 $ & $0.54 \mp 0.01 $ & $\mathbf{0.60} \mp 0.01$& $\underline{0.59} \mp 0.02$ & $0.57 \mp 0.01$\\ 
                                       & ACC & $\mathbf{0.55} \mp 0.02 $ & $0.45 \mp 0.03 $ & $0.48 \mp 0.01 $       &  $\underline{0.52} \mp 0.04$ & $0.48 \mp 0.01$\\ 
      \midrule
        \multirow{2}{*}{Letter}        & ARI & $\mathbf{0.30} \mp 0.01 $ & $0.07 \mp 0.03 $ & $\underline{0.23} \mp 0.05$ & $\mathbf{0.30} \mp 0.01$ & $0.20 \mp 0.00$\\
                                       & NMI & $\underline{0.56} \mp 0.01 $ & $0.23 \mp 0.05 $ & $0.47 \mp 0.06 $ & $\mathbf{0.61} \mp 0.00$ & $0.52 \mp 0.00$\\ 
                                       & ACC & $\mathbf{0.41} \mp 0.01$ & $0.14 \mp 0.04 $ & $0.31 \mp 0.06 $ & $\underline{0.33} \mp 0.01$ & $0.24 \mp 0.01$\\ 
      \midrule
        \multirow{2}{*}{Balanced}      & ARI & $\mathbf{0.35} \mp 0.00 $ & $0.02 \mp 0.01 $ & $0.05 \mp 0.02 $ & $\underline{0.35} \mp 0.01$ & $0.22 \mp 0.00$\\
                                       & NMI & $\underline{0.59} \mp 0.00 $ & $0.17 \mp 0.06 $ & $0.31 \mp 0.04 $ & $\mathbf{0.63} \mp 0.00$ & $0.54 \mp 0.00$\\ 
                                       & ACC & $\mathbf{0.46} \mp 0.01 $ & $0.07 \mp 0.02 $ & $0.10 \mp 0.02 $ & $\underline{0.44} \mp 0.02$ & $0.30 \mp 0.01$\\ 
      \midrule
        \multirow{2}{*}{Digits}        & ARI & $\underline{0.55} \mp 0.01 $ & $0.28 \mp 0.14 $ & $\mathbf{0.74} \mp 0.05 $ & $0.46 \mp 0.02$ & $0.36 \mp 0.01$\\
                                       & NMI & $\underline{0.71} \mp 0.01 $ & $0.51 \mp 0.14 $ & $\mathbf{0.79} \mp 0.02 $ & $\underline{0.71} \mp 0.01$ & $0.63 \mp 0.00$\\
                                       & ACC & $\underline{0.58} \mp 0.00 $ & $0.38 \mp 0.09 $ & $\mathbf{0.81} \mp 0.05 $ & $0.44 \mp 0.03$ & $0.34 \mp 0.01$\\ 
      \midrule
        \multirow{2}{*}{UrbanGB}       & ARI & $\underline{0.63} \mp 0.01 $ & $0.12 \mp 0.05 $ & $0.09 \mp 0.00 $ & $\mathbf{0.67} \mp 0.05$ & $0.49 \mp 0.06$\\
                                       & NMI & $0.68 \mp 0.01 $ & $0.21 \mp 0.07 $ & $0.24 \mp 0.00 $ & $\underline{0.77} \mp 0.01$ & $\mathbf{0.81} \mp 0.01$\\
                                       & ACC & $0.45 \mp 0.01 $ & $0.29 \mp 0.02 $ & $0.29 \mp 0.00 $ & $\underline{0.53} \mp 0.02$ & $\mathbf{0.54} \mp 0.02$\\ 
     \bottomrule    
 \end{tabular}}
\caption{The mean and the standard deviation of the three clustering metrics, ARI, NMI, and ACC, over $10$ runs on different datasets. The best result within each row is marked as bold, and the runner-up is underlined.
}
\label{clustresults}
\end{table}

\begin{figure}[t!]
    \centering
    \includegraphics[scale=0.27]{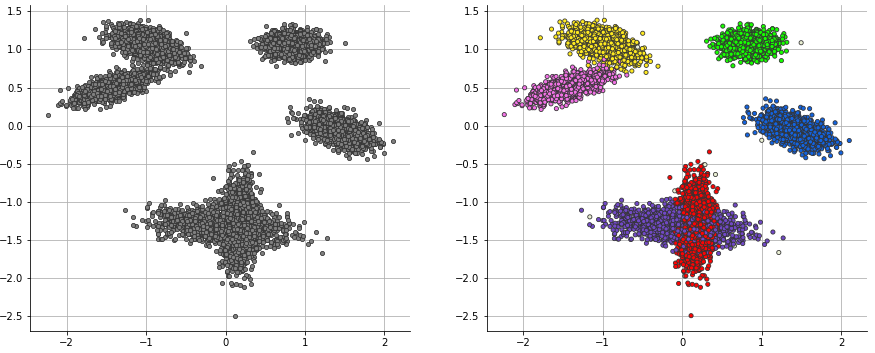}
    \caption{Unlabeled (left) and labeled data (right), after
 $100$ iterations of DisCGS on a synthetic dataset with overlapped clusters, ARI $=0.85$, NMI $=0.89$, and ACC $=0.91$. The number of estimated clusters is $7$.}
    \label{fig:synth10k}
\end{figure}

Figure \ref{fig:synth10k} illustrates the estimated clusters obtained by our distributed approach on the synthetic 10K dataset. This figure represents the best partition achieved over the ten trials. The purpose of using this dataset is to evaluate DisCGS's performance when confronted with complex datasets that exhibit overlapping clusters. The results demonstrate that our method performs remarkably well even when dealing with such datasets. 
DisCGS achieved an ARI, NMI, and ACC score of 0.85, 0.89, and 0.91, respectively. It estimated seven clusters, while the number of clusters is six.

\subsection{Convergence}
In this experiment, we examine the convergence of both likelihood and ARI score of three methods: DisCGS, CGS (Collapsed Gibbs Sampler), and SubC. These evaluations are performed on three datasets: Synthetic 100K, Fashion-mnist, and Balanced. It's important to note that only CGS and SubC share the same model assumption as DisCGS, resulting in a comparable likelihood. 

\begin{figure}[h!]
    \centering
    \includegraphics[scale=0.425]{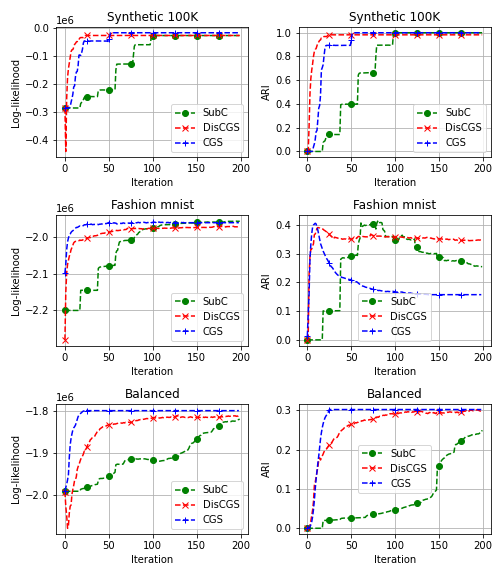}
    \caption{Log-likelihood and ARI score every iteration}
    \label{fig:likelihood}
\end{figure}

Figure \ref{fig:likelihood} illustrates the evolution of the log-likelihood and ARI score at each iteration. We observe that our algorithm converges almost at a similar rate as the centralized CGS algorithm and is much faster than SubC, which takes more iterations to converge. This is because our algorithm is able to discover new local clusters inside each worker. Whereas, in SubC, the number of clusters is fixed when performing the restricted Gibbs sampler in each worker, and new components are only discovered at the master level during the split step. Thus, more iterations are required to generate enough components to model the data. This phenomenon is also observed in \cite{Wang2017ScalableEO}. Overall, our algorithm converges really fast and maintains a stable ARI score over the iterations. Whereas the CGS and SubC may downgrade their ARI score after some iterations, as can be observed for the Fashion-Mnist dataset.   
\subsection{Comparison of the distributed and centralized collapsed Gibbs sampler}
In this experience, we compare the execution time and clustering performance of the distributed collapsed Gibbs sampler (DisCGS) and the centralized collapsed Gibbs sampler (CGS) from \cite{10.2307/1390653}, we execute both algorithms on synthetic datasets of different sizes (from $n=20$K to $n=100$K) generated from $K=10$ Gaussian components of dimension $2$. The centralized version is too slow; running over $100$K observations would take too much time. We use 32 cores for this experiment.
\begin{figure}[htbp]
    \centering
    \includegraphics[scale=0.35]{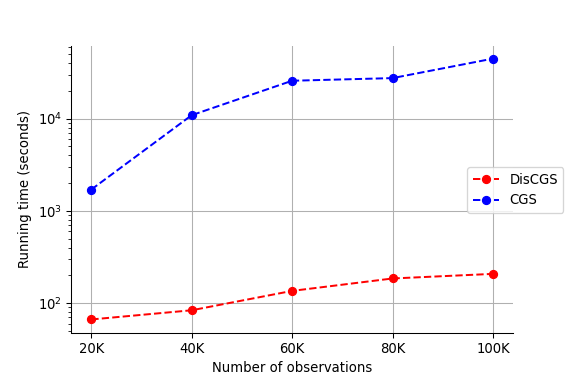}
    \caption{Running time (in logarithmic scale) for $100$ iterations of DisCGS and CGS inference for DPMM on synthetic datasets of different sizes.}
    \label{fig:DisVsCent}
\end{figure}

Figure \ref{fig:DisVsCent} illustrates, using a logarithmic scale, the execution time for $M=100$ iterations of both the distributed and centralized inference methods. The results show that our distributed algorithm significantly outperforms the centralized approach. For instance, when considering $100$K data points, the centralized algorithm takes approximately $12$ hours to complete $100$ iterations, whereas our approach achieves the same number of iterations in just $3$ minutes, reducing the execution time by a factor of $200$. Table \ref{table:DisVsCent} presents the clustering metrics (ARI, NMI, and ACC) obtained by each algorithm on each dataset. The results indicate that our approach consistently achieves high scores and outperforms the centralized algorithm in almost all cases. We observe that using the dataset of size $40$K, the centralized algorithm obtained slightly higher ARI and NMI scores than the distributed algorithm. This is not surprising since both approaches sample the memberships from an approximation of the posterior distribution, resulting in noisy inferred partitions. These samples can be aggregated after a given number of burn-in iterations with a consensus partition estimation. Overall, the findings confirm that the distributed inference does not compromise the clustering performance while considerably reducing the execution time.

\begin{table}[htbp]
\resizebox{\columnwidth}{!}{%
  \begin{tabular}{lcccccc}
    \toprule
    \multirow{2}{*}{Dataset size} &
      \multicolumn{2}{c}{ARI} &
      \multicolumn{2}{c}{NMI} &
      \multicolumn{2}{c}{ACC} \\  
      & {Dis.} & {Cen.} & {Dis.} & {Cen.} & {Dis.} & {Cen.}\\
      \midrule
    20K  & \textbf{0.99} & 0.89 & \textbf{0.99} & 0.96 & \textbf{0.99} & 0.89 \\
    40K  & 0.96 & \textbf{0.99} & 0.97 & \textbf{0.99} & \textbf{0.99} & 0.97 \\
    60K  & \textbf{0.91} & 0.89 & 0.92 & \textbf{0.96} & \textbf{0.92} & 0.89 \\
    80K  & \textbf{0.94} & 0.89 & \textbf{0.96} & \textbf{0.96} & \textbf{0.91} & 0.89 \\
    100K & \textbf{0.91} & 0.89 & \textbf{0.94} & 0.89 & \textbf{0.91} & 0.89 \\    
    \bottomrule
  \end{tabular}}
          \caption{Clustering metrics ARI, NMI, and ACC obtained by the distributed (Dis.) and centralized (Cen.) inference on synthetic datasets of different sizes.}
          \label{table:DisVsCent}
\end{table}
\subsection{Distributed algorithm scale-up}

In this experiment, we use $n=10^6$ data points generated from $K=10$ components of 
two-dimensional Gaussian. We run our distributed inference several times by increasing the number of cores from $8$ to $48$. 

\begin{figure}[htbp]
\centering
    \includegraphics[scale=0.35]{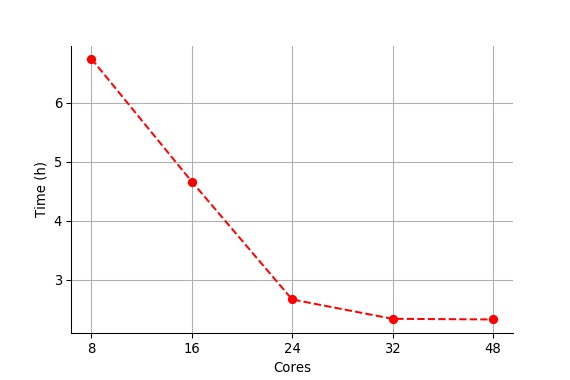}        
    \caption{Running time (hours) as a function of the number of cores of  DisCGS on $10^6$ data points.}
    \label{fig:Cores}
\end{figure}

\begin{figure}[htbp]
    \centering
    \includegraphics[scale=0.25]{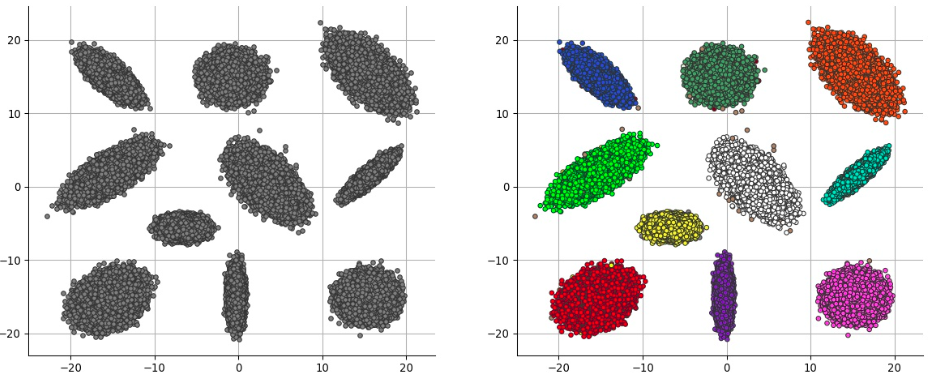}
    \caption{Unlabeled data (left) and labeled data (right), after $M=100$
iterations of DisCGS on $10^6$ data points, using $32$ cores, ARI $=0.98$, NMI $=0.98$, and ACC $=0.98$. The number of inferred clusters is $14$.}
    \label{fig:Labels}
\end{figure}

Figure \ref{fig:Cores} represents the running time as a function of the number of cores. We observe that the running time decreases when the number of cores increases, showing that our algorithm scales efficiently with the number of workers. 
Figure \ref{fig:Labels} shows the clusters (labels) inferred by our approach on $10^6$ data points distributed across $32$ cores. As depicted, our approach successfully identifies meaningful and coherent clusters, resulting in high ARI, NMI, and ACC scores of approximately $0.98$. Similar results were observed when distributing the data on different numbers of cores. Additionally, it has inferred $14$ clusters while the ground truth is $10$. However, only $10$ of them are significant clusters, and the $4$ others are only "outliers". Moreover, the clustering performance and the number of clusters depend on the auto-encoder. Also, the number of clusters might be influenced by the concentration parameter $\alpha$ \cite{murphy2013machine}. 

\section{Conclusion and perspectives}

This article presents a novel distributed MCMC inference method, called DisCGS, for DPMMs. DPMMs are highly useful for clustering problems, especially when the number of clusters is unknown. However, the inference process of DPMMs tends to become significantly slow as the dataset size increases. To overcome this limitation, our proposed DisCGS is specifically designed to handle distributed data across independent and heterogeneous machines,  making it suitable for horizontal federated learning scenarios, i.e., when the workers are the different components of the edge network. The experimental results showed highly promising outcomes. The proposed method significantly reduces the inference time while maintaining accurate results. 
Our ongoing research exploits the DisCGS approach to massively distribute the  Non-Parametric Latent Block Model (NPLBM) \cite{Meeds07nonparametricbayesian} and the multiple Coclustering model. 

\section{Acknowledgments}
This work has been supported by the Paris Île-de-France Région in the framework of DIM AI4IDF.
\bibliographystyle{abbrv}
\bibliography{bibliography}

\end{document}